\documentclass[preprint,12pt]{elsarticle}

\usepackage{amssymb}
\usepackage{amsthm}
\usepackage{bbold}
\usepackage{adjustbox}
\usepackage{graphicx}
\usepackage{CJKutf8}
\usepackage{color}

\journal{Neurocomputing}

\begin{document}

\begin{frontmatter}



\title{JCSE: Contrastive Learning of Japanese Sentence Embeddings and Its Applications}


\author[label1]{Zihao Chen}
\author[label2]{Hisashi Handa}
\author[label2]{Kimiaki Shirahama}

\affiliation[label1]{organization={Graduate School of Science and Engineering, Kindai University},%Department and Organization
            addressline={3-4-1, Kowakae, Higashiosaka}, 
            city={Osaka},
            postcode={577-8502},
            state={},
            country={Japan}}

\affiliation[label2]{organization={Faculty of Informatics, Kindai University},%Department and Organization
            addressline={3-4-1, Kowakae, Higashiosaka}, 
            city={Osaka},
            postcode={577-8502},
            state={},
            country={Japan}}

\begin{abstract}
Contrastive learning is widely used for sentence representation learning. Despite this prevalence, most studies have focused exclusively on English and few concern domain adaptation for domain-specific downstream tasks, especially for low-resource languages like Japanese, which are characterized by insufficient target domain data and the lack of a proper training strategy. To overcome this, we propose a novel Japanese sentence representation framework, JCSE (derived from ``Contrastive learning of Sentence Embeddings for Japanese''), that creates training data by generating sentences and synthesizing them with sentences available in a target domain. Specifically, a pre-trained data generator is finetuned to a target domain using our collected corpus. It is then used to generate contradictory sentence pairs that are used in contrastive learning for adapting a Japanese language model to a specific task in the target domain.

% we focus on domain adaptation where a general Japanese language model is adapted to specific downstream tasks.
% Afterward, we perform contrastive learning with a two-stage training strategy, first incorporating the synthesized triple sentence pairs into \textcolor{red}{Japanese language models (suddenly appear)}, then finetuning them on \textcolor{red}{JSNLI} data \textcolor{red}{Is this the target data?}.

Another problem of Japanese sentence representation learning is the difficulty of evaluating existing embedding methods due to the lack of benchmark datasets. Thus, we establish a comprehensive Japanese Semantic Textual Similarity (STS) benchmark on which various embedding models are evaluated. Based on this benchmark result, multiple embedding methods are chosen and compared with JCSE on two domain-specific tasks, STS in a clinical domain and information retrieval in an educational domain. The results show that JCSE achieves significant performance improvement surpassing direct transfer and other training strategies. This empirically demonstrates JCSE's effectiveness and practicability for downstream tasks of a low-resource language.

% This prevents us from choosing embedding models that are compared to the one trained by JCSE. 
% \textcolor{red}{domain adaptation (suddenly appears)} for specific domain downstream tasks. In particular, there is no such research \textcolor{red}{in Japanese} due to insufficient target domain data and the lack of a proper \textcolor{red}{sentence embedding (suddenly appear)} training strategy.

\end{abstract}

\begin{graphicalabstract}
\end{graphicalabstract}

\begin{highlights}
\item We propose JCSE that is a novel effective and efficient Japanese sentence representation learning framework for domain adaptation on Japanese domain-specific downstream tasks.
\item We construct and release a comprehensive Japanese STS (Semantic Textual Similarity) benchmark. To our best knowledge, we are the first to evaluate multiple Japanese sentence embedding models on this comprehensive benchmark.
\item We propose a new human-annotated QAbot dataset for an information retrieval task in a Japanese educational domain.
\end{highlights}


\end{frontmatter}


\section{Introduction}
\label{sec:intro}

Sentence representation learning is a fundamental problem in natural language processing (NLP). It learns to encode sentences into compact, dense vectors, called embeddings, which can be broadly useful for a wide range of tasks, such as  Semantic Textual Similarity (STS)~\cite{ref1}, language understanding and evaluation~\cite{ref2,ref3} and information retrieval~\cite{ref4,ref5}.

Recent work has shown that finetuning pre-trained language models with contrastive learning is the most effective and efficient for sentence representation learning. The instance of contrastive learning makes embeddings of semantically similar pairs close to each other while pushing dissimilar ones apart~\cite{ref6}. In contrastive learning, previous methods \cite{ref7,ref8,ref9} use multiple ways to construct sentence pairs for supervised or unsupervised training. Specifically, supervised contrastive methods \cite{ref7,ref10} usually create  sentence pairs using several publicly available Natural language inference (NLI) datasets, like Stanford NLI (SNLI) dataset \cite{ref11} and MultiGenre NLI (MNLI) dataset \cite{ref12}. Meanwhile, unsupervised contrastive methods \cite{ref7,ref13,ref14} leverage unlabeled sentences from Wikipedia or any other public resources. In general, supervised contrastive learning methods outperform unsupervised ones although the former need multi-source large-scale labeled training data. Usually, previous research follows the standard evaluation protocol that uses the SentEval toolkit \cite{ref15} on the standard semantic textual similarity (STS) benchmark for evaluation purposes. Accordingly, several benchmark datasets for STS and NLI have been released by targeting at English.

When it comes to Japanese, things are different. Although JSNLI~\cite{ref22} and JSICK~\cite{ref23} datasets are publicly available, STS12-16~\cite{ref16,ref17,ref18,ref19,ref20}, STS-B~\cite{ref21} and 7 transfer tasks from SentEval~\cite{ref15}, and MNLI datasets~\cite{ref12} in Japanese do not exist. Here, JSNLI dataset is translated from the SNLI dataset using machine translation, and JSICK dataset is made manually based on the SICK~\cite{ref24} dataset. We believe that the lack of publicly available benchmark datasets for Japanese STS has led to the lack of interest in building Japanese sentence representation models and devising applications based on them. Motivated by this, we believe it is time to construct and release a comprehensive Japanese STS benchmark. In addition, we publish various Japanese sentence embedding models trained mainly by contrastive learning and propose a novel effective method as our core work of this paper for the domain-specific downstream tasks.

In this context, we propose an efficient Japanese sentence representation learning framework that leverages data generation to perform contrastive learning on domain-specific downstream tasks. This framework is named JCSE that is derived from ``Contrastive learning of Sentence Embeddings for Japanese''. JCSE is built upon pre-trained transformer language models~\cite{ref25,ref26}. We first collect a domain-specific raw corpus in a target domain. Afterward, we finetune a pre-trained data generator (in our case T5~\cite{ref26}) on our collected domain-specific corpus using unsupervised fill-in-the-blank-style denoising training. It is then used to construct negative (contradictory) pairs for input sentences. In previous research~\cite{ref23}, Japanese language models predict entailment labels without considering word order or case particles. The experimental settings for whether the existence of case particles can change entailment labels~\cite{ref23} inspire us to investigate which word types in Japanese are the most relevant for different Japanese language models. The results of our relevant content words experiments~\cite{ref27} show that nouns are the most relevant content words in a Japanese sentence. It means that negative pairs for input sentences can be constructed by substituting nouns in them using the finetuned data generator. Referring to SimCSE \cite{ref7}, we construct positive pairs by simply using different dropout masks. In this way, sentence pairs are synthesized for contrastive learning. Finally, we apply a two-stage contrastive learning recipe that finetune a Japanese language model first on synthesized sentence pairs and then on JSNLI.

To evaluate our method JCSE compared with various existing embedding methods, we choose multiple proper Japanese sentence embedding models based on the results of our established Japanese STS benchmark. On the STS benchmark, Japanese sentence embedding models yield similar results to the case of English, that is, applying contrastive learning to sentence embedding models leads to performance improvements. Concern about our Japanese STS benchmark, as described above, STS12-16 and STS-B in Japanese still do not exist. Referring to previous research~\cite{ref22,ref28}, machine translation is used to translate English STS12-16 and STS-B datasets into Japanese. Combining the translated datasets with existing JSICK~\cite{ref23} and Japanese STS (JSTS) datasets included in the JGLUE~\cite{ref53}, we construct and release this comprehensive version of the Japanese STS benchmark\footnote[1]{https://github.com/mu-kindai/JCSE}.

In parallel, with the direct transfer of Japanese sentence embedding models to the two domain-specific downstream tasks, contrastive learning leads to better performance than previous embedding methods, but those improvements are diminutive.
Through further domain adaptation with our method, JCSE, significantly improves the performance of these two domain-specific downstream tasks, evaluating the performance on Japanese Clinical STS (JACSTS) dataset \cite{ref29} and QAbot dataset. JACSTS dataset is created for an STS task on Japanese sentences in a clinical domain. QAbot dataset is collected by us for an information retrieval task in an educational domain of computer science. The results show JCSE achieves competitive performance after the first finetuning using synthesized sentence pairs, and attains further improvement after the second finetuning of semi-supervised training on JSNLI. Surprisingly, we find that directly integrating our collected domain-specific corpus into the Japanese sentence embedding models using previous state-of-the-art unsupervised contrastive learning methods can also significantly improve the performance of the domain-specific downstream tasks.

The main contributions of this paper are summarized as follows:
\begin{itemize}
    \item We propose JCSE that is a novel effective and efficient Japanese sentence representation learning framework for domain adaptation on Japanese domain-specific downstream tasks.
    \item We construct and release a comprehensive Japanese STS benchmark. To our best knowledge, we are the first to evaluate multiple Japanese sentence embedding models on this comprehensive benchmark. % We train various Japanese sentence embedding models and perfrom evaluations using the SentEval toolkit on the Japanese STS benchmark.
    \item We propose a new human-annotated QAbot dataset for an information retrieval task in a Japanese educational domain.
\end{itemize}

\section{Background and Related Work}

\subsection{Contrastive Learning}

Contrastive learning aims to learn effective representation by pulling semantically close neighbors together and pushing apart non-neighbors \cite{ref30}. A training set of contrastive learning is defined by paired examples each of which is denoted by $D=\{(v_{i},v_{i}^{+})\}$. Here, $v_i$ is the embedding of the $i$th sentence and $v_i^{+}$ is the one of a sentence semantically related to the $i$th sentence. During training, $v_{i}^+$ is considered a positive example for $v_{i}$ and all other examples in a batch are considered negatives \cite{ref31,ref32}. For a mini-batch of $N$ pairs, the contrastive loss function is defined as:

\begin{equation} \label{eq1}
    {\mathcal{L} }  =  \frac{e^{\mathrm{sim} (v_{i},v_{i}^+)/\tau } }{ {\textstyle \sum_{j=1}^{N}} {e^{\mathrm{sim} (v_{i},v_{j}^+)/\tau}}},
\end{equation}
where $\tau$ is a temperature hyperparameter, $\mathrm{sim} (v_{1}, v_{2})$ is the cosine similarity $\frac{{v_{1}}^\top v_{2}}{\left \|v_{1}  \right \| \cdot \left \|v_{2}  \right \| } $. When additional negatives $v_{j}^-$ are provided for $v_i$, the contrastive loss can be enhanced as:

\begin{equation} \label{eq2}
    {\mathcal{L} }  = \frac{e^{\mathrm{sim} (v_{i},v_{i}^+)/\tau } }{ {\textstyle \sum_{j=1}^{N}} ({e^{\mathrm{sim} (v_{i},v_{j}^+)/\tau}}+{e^{\mathrm{sim} (v_{i},v_{j}^-)/\tau}})}.
\end{equation}

SimCSE~\cite{ref7} applies different hard negatives from other in-batch negatives, extends Eq.~\ref{eq2} to incorporate weighting of different negatives:

\begin{equation} \label{eq3}
    {\mathcal{L} }  = \frac{e^{\mathrm{sim} (v_{i},v_{i}^+)/\tau } }{ {\textstyle \sum_{j=1}^{N}} ({e^{\mathrm{sim} (v_{i},v_{j}^+)/\tau}}+\alpha ^{\mathbb{1} _{i}^{j} } {e^{\mathrm{sim} (v_{i},v_{j}^-)/\tau}})},
\end{equation}
where ${{\mathbb{1} _{i}^{j}}\in \left \{ 0,1 \right \} } $  is an indicator that equals 1 if and only if $i=j$, and $\alpha$ is a weighting hyperparameter.

One crucial question in contrastive learning is how to construct $(v_{i},v_{i}^+)$ pairs. Some research \cite{ref8,ref9,ref33} apply multiple augmentation techniques such as word deletion, reordering, cutoff, and substitution. Kim \textit{et al.}~\cite{ref34} directly use the hidden representation of BERT as well as its final sentence embeddings. SimCSE~\cite{ref7} simply feeds the same input to the encoder twice and constructs positive pairs using different dropout masks. This simple method has proved to be more effective than other strategies. Previous methods mainly focus on constructing positive pairs while negative pairs are generated through random sampling. Several works~\cite{ref35,ref36} in the computer vision community show that hard negative examples are essential for contrastive learning. Robinson \textit{et al.}~\cite{ref37} use an importance sampling method to select hard negative examples. For NLP, Kalantidis \textit{et al.}~\cite{ref38} improve MoCo~\cite{ref39} by generating hard negative examples through mixing positive and negative examples in the memory bank which keeps a queue with features of the last few batches. MixCSE~\cite{ref13} constructs artificial hard negative features via mixing both positive and negative samples for unsupervised sentence representation learning. While supervised sentence representation learning relies on large-scale human-annotated sentence pairs, we often face with the scarcity of human-annotated datasets, especially low-resource languages like Japanese.

\subsection{Domain Adaption for downstream task}

Domain Adaptation is a fundamental challenge in NLP. The most common domain adaption technique for a pre-trained language model is domain-adaptive pre-training which continues pre-taining on in-domain data. Recent work~\cite{ref40} has shown empirically that domain adaptation improves a language model's performance on a  target domain. Specifically, the researchers in \cite{ref40} experiment with eight classification tasks ranging across four domains (biomedical and computer science publications, news, and reviews). The results show that continued pre-training of an already powerful language model on the domain consistently improves performance on tasks in a target domain. BioBERT~\cite{ref41} shows that pre-training BERT on biomedical corpora largely improves its performance on downstream tasks of biomedical text mining. Meanwhile, Guo \textit{et al.}~\cite{ref42} and Ma \textit{et al.}~\cite{ref43} show that neural retrieval models trained on a general domain often do not transfer well, especially for specialized domains.

There is no labeled training data in many target domains. To overcome this, leveraging generative models to produce synthetic data has been previously explored. Du \textit{et al.}~\cite{ref44} first apply a seq2seq model for automatic question generation from text passages in a reading comprehension task. Puri \textit{et al.}~\cite{ref45} first demonstrate that a QA model trained on purely synthetic questions and answers can outperform models trained on human-labeled data on SQuAD1.1. Ma \textit{et al.}~\cite{ref43} and Liang \textit{et al.}~\cite{ref46} use synthetic queries for domain adaptive neural retrieval. Yue~\textit{et al.}~\cite{ref47} leverage question generation to synthesize QA pairs on clinical contexts and boost QA models without requiring manual annotations. Another research \cite{ref48} trains a QA system on both source data and generated data from a target domain with a contrastive adaptation loss. However, to our best knowledge, there is no recent research that addresses domain adaptation in a domain-specific task involving Japanese sentences.

\subsection{Non-English data construction}

With the development of multilingual pre-trained language models, general language understanding frameworks for languages other than English have been created, and NLI datasets have been multilingualized. Conneau \textit{et al.}~\cite{ref49} provided a cross-lingual NLI (XNLI) dataset by translating MNLI \cite{ref12} into 15 languages, including languages with few language resources such as Urdu and Swahili. However, Japanese is not included in it.

As examples of non-English datasets, OCNLI \cite{ref50} is a Chinese NLI dataset built from original Chinese multi-genre resources. Ham \textit{et al.}~\cite{ref28} create KorNLI and KorSTS datasets by translating English NLI and STS-B \cite{ref21} datasets into Korean. There are also attempts to translate the SICK dataset \cite{ref24} into Dutch \cite{ref51} and Portuguese \cite{ref52}.

Regarding Japanese, a Japanese SNLI (JSNLI) dataset \cite{ref22} is constructed by employing machine translation to translate the English SNLI dataset into Japanese and automatically filtering out unnatural sentences. JSICK \cite{ref23} dataset is manually translated from the English SICK dataset \cite{ref24}. Yahoo Japan constructs a Japanese GLUE benchmark \cite{ref53} from scratch without translation due to the cultural/social discrepancy between English and Japanese. Although employing machine translation incurs several possibilities to produce unnatural sentences, it is more cost-efficient than manual translation and reconstruction from scratch.

\section{Contrastive Learning of Sentence Embeddings for Japanese}

\begin{figure}[htp]
    \centering
    \includegraphics[width=\textwidth]{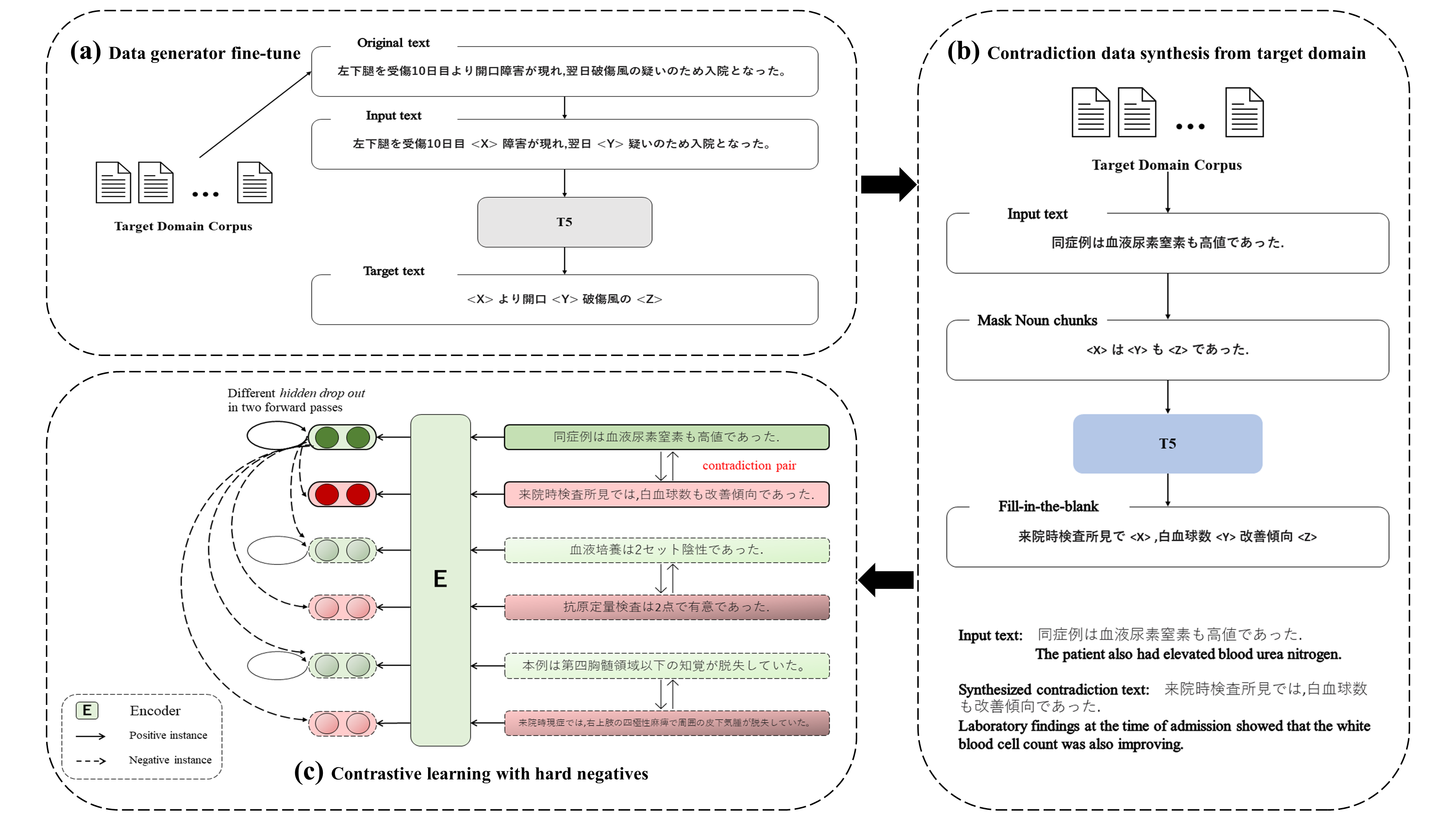}
    \caption{An overview of JCSE.}
    \label{fig1}
\end{figure}

% In this section, we describe our method, JCSE, a Japanese sentence representation learning framework for Japanese domain-specific downstream tasks as illustrated in Fig. \ref{fig1}.

Fig.~\ref{fig1} illustrates an overview of JCSE. For a target domain, we collect a raw corpus from public resources and use GINZA\footnote[2]{https://github.com/megagonlabs/ginza}, an open-source Japanese NLP library, to divide the corpus into sentences. Next, we carry out text normalization like removing markup and non-text content, then drop sentences whose length is less than five after the tokenization by GINZA. A pre-trained T5 model is finetuned. Afterward, unlabeled sentences in the target domain are fed into the finetuned T5 model to generate negative pairs. Then, positive pairs are defined by simply using different dropout masks, and the combination of positive and negative pairs forms synthesized input pairs. Finally, we train a sentence embedding model using contrastive learning based on synthesized pairs and human-annotated sentence pairs, here is the JSNLI dataset. More details of the above-mentioned processes are described below.

\subsection{Collecting target domain corpora}

We collect Japanese corpora for two target domains, a clinical domain and an educational computer science domain. The scarcity of domain-specific datasets in low-resource languages such as Japanese is immense, especially human-annotated supervised data. We collect the raw target domain corpora to alleviate the data scarcity.

For the Japanese clinical domain, the reason for few resources is due to the data privacy, which prohibits public sharing of medical data. We collect data from two available sources, Japanese Case Reports (CR) and Japanese NTCIR-13 MedWeb dataset combined in the NTCIR-13 MedWeb task published at the NTCIR-13 conference\footnote[3]{http://mednlp.jp/medweb/NTCIR-13/}. More specifically, the CR dataset is created by 
extracting 200 case reports from CiNii\footnote[4]{https://cir.nii.ac.jp/}, a Japanese database containing research publications. The Japanese NTCIR-13 MedWeb dataset is a part of the NTCIR-13 Medical NLP for Web Document task, which provides $2560$ Twitter-like message texts. We combine these two datasets, drop duplicate ones and then conduct text normalization to remove markup, special symbols and non-relative text contents in order to construct a clean Japanese clinical domain corpus.

% The dataset consists of 2560 disease/symptom-related tweets in Japanese in total. 

The Japanese educational computer science domain is very special because of its interdisciplinary characteristic to integrate education and computer science knowledge. We leverage our university's syllabus data and teaching materials published in Google Classroom of courses in the department of informatics. In addition, private question-answer data are collected by running a QAbot on Slack\footnote[5]{https://slack.com/} which is a popular instant messaging software and utilized in different courses of our department. We incorporate the data from these three resources (Google Classroom, syllabus and QAbot data) and do text normalization to get a clean Japanese educational computer science domain corpus. We publicly share this unique private corpus. All the corpora we collected are available here\footnote[6]{https://github.com/mu-kindai/JCSE}.

\subsection{Data generation}

We choose T5 \cite{ref26} as our data generation model. More specifically, we apply the Japanese T5 Version 1.1 model which is an improved version of the original T5 model. This new version released by Google is built by unsupervised pre-training and has not been fine-tuned on any downstream task. We formulate data generation as a fill-in-the-blank task where the T5 model predicts missing words within a corrupted piece of text in a text-to-text format.

As described in Section~\ref{sec:intro}, we find that nouns are the most relevant content words in Japanese sentences. Hence, for the target domain unlabeled sentences, we first perform part-of-speech (POS) tagging for every sentence using GINZA and spaCy \cite{ref54}. The POS-tags labels are determined using spaCy \cite{ref54}. Then we mask all noun tokens, more precisely mask noun chunks rather than noun tokens to generate more natural data. Each corrupted noun chunk is replaced by a sentinel token that is unique over the example. The sentinel token is assigned a token ID that is unique to the sentence. The sentinel IDs are special tokens that are added to the T5 model's vocabulary and do not correspond to any word piece \cite{ref26}. We formulate our data generation as a fill-in-the-blank task where  the input is a sentence corrupted with multiple sentinel tokens and the target output is the relevant text in the missing location. It is illustrated in Figure~\ref{fig1} (b) where \textless X\textgreater, \textless Y\textgreater \ and \textless Z\textgreater \ in the box annotated with ``Mask Noun chunks'' are filled with terms in the box with ``Fill-in-the-black''.

A pre-trained T5 model already has the ability to carry out the fill-in-the-blank task without finetuning.  However, it is pre-trained on general domain corpora without considering sufficient knowledge of specific domains such as clinical and computer science domains. Also, the word distribution of general and specific domain corpora are quite different, so there is no ability for the pre-trained T5 model to generate proper domain-specific terms. Hence, it is essential to finetune the pre-trained T5 model on unlabeled data in the target domain using unsupervised denoising training. Referring to T5 official training settings \cite{ref26}, we randomly mask 15 percent of tokens and keep the average length of spans as 3. It is illustrated in Figure~\ref{fig1} (a). The fine-tuned T5 model is used as our data generator.

\subsection{Contrastive learning based on hard negatives}

% Contrastive learning has been widely used in sentence representation learning, achieving state-of-the-art performance on the STS benchmark and downstream tasks \cite{ref7,ref10,ref13,ref14}. In this work, 

Inspired by SimCSE \cite{ref7}, contrastive learning based on hard negatives is performed as illustrated in Figure~\ref{fig1} (c). During training, we regard a training example as a triplet $\left \{ v,v^{+},v^{-}  \right \}$ where $v$ and $v^{+}$ from an entailment/positive pair, $v$ and $v^{-}$ constitute a contradictory/negative pair. We also apply in-batch negative sampling where all other examples in a batch are considered as negatives. For a mini-batch of $N$ triple pairs, the hard negatives based contrastive learning objective is defined as Eq. \ref{eq4} which incorporates weighting of different negatives:

\begin{equation} \label{eq4}
-\mathrm{log}\frac{e^{\mathrm{sim} (v_{i},v_{i}^+)/\tau } }{ {\textstyle \sum_{j=1}^{N}} ({e^{\mathrm{sim} (v_{i},v_{j}^+)/\tau}}+\alpha ^{\mathbb{1} _{i}^{j} } {e^{\mathrm{sim} (v_{i},v_{j}^-)/\tau}})},
\end{equation}
and minimize the following contrastive learning loss during training:

\begin{equation} \label{eq5}
    {\mathcal{L} }  =  \frac{e^{\mathrm{sim} (v_{i},v_{i}^+)/\tau } }{ {\textstyle \sum_{j=1}^{N}} ({e^{\mathrm{sim} (v_{i},v_{j}^+)/\tau}}+\alpha ^{\mathbb{1} _{i}^{j} } {e^{\mathrm{sim} (v_{i},v_{j}^-)/\tau}})},
\end{equation}
where $i$ and $j$ are the sentence indices in the mini-batch, $\tau$ is the temperature hyperparameter, $\mathrm{sim} (\cdot, \cdot )$ is a function to compute cosine similarities, $\alpha$ is the weighting hyperparameter, and ${{\mathbb{1} _{i}^{j}}\in \left \{ 0,1 \right \} } $  is an indicator that equals 1 if and only if $i=j$.

\section{Experiments}
In this section, we describe the experiments for this paper. To evaluate the performance of existing sentence representation methods and choose the ones compared to JCSE, first we establish a Japanese STS benchmark and evaluate various Japanese sentence embedding models on it. Then we do the relevant content words experiments and find that nouns are the most relevant content words for Japanese language models. Afterward, we generate contradictory sentences using finetuned 
 data generator and conduct domain adaptation for downstream tasks using contrastive learning with a two-stage training recipe. Last, we evaluate JCSE compared with direct transfer and other training strategies on two domain-specific downstream tasks. More experiment details are described following.

\subsection{Japanese STS benchmark Construction}
We release a Japanese STS benchmark in our work. We combine the published JSICK \cite{ref23} and JSTS \cite{ref53} datasets with translated STS12-16 and STS-B datasets using machine translation. The translated datasets are derived from original English datasets. We apply a similar translation strategy by referring  to the previous methods~\cite{ref22,ref28}.

First, we translate all of the STS12-16 and STS-B datasets into Japanese using Google translate API\footnote[7]{https://cloud.google.com/translate/} and DeepL translate API\footnote[8]{https://www.deepl.com/en/pro-api}. Then, the translated sentences are 
automatically filtered based on their BLEU scores \cite{ref55}. More specifically, we translate each of the translated Japanese datasets back into English using the same machine translator, then we compute the BLEU1 score between the 
original and translated English sentences. We find that most of the BLEU scores are 0 when using $N>1$ of $N$-gram, so we just apply the BLEU1 score. Unlike the previous research \cite{ref22,ref28}, we do not translate sentences in the development and test sets using human translation or post-edit translation results by  professional translators due to cost efficiency.

Last, to ensure the quality and decide on the translation data translated from Google or DeepL translator, we conduct an experiment by referring to the previous method \cite{ref22} that trains a linear regression model on different translated data defined by diverse BLEU1 score thresholds. Detailedly, a linear model is built by adding a linear layer on the top layer's $CLS$ token of a pre-trained BERT model. It is then trained on the translated STS datasets to evaluate the performance of each translator on the JSTS, the JSTS dataset is manually created so can be seen as gold labels. All models are trained in 3 epochs and the same settings. The training data defined by varied BLEU1 score thresholds have different data qualities and sizes that influence the model performance of the linear regression task. The results are shown in Table \ref{tab1}. Based on the results and concerns about the balance between the 
data size and model performance, we choose all the data translated from the DeepL translator. Translated sentences whose BLEU1 score is 0 are filtered out automatically. We check these dropped-out data and discover issues such as gibberish and ambiguity. We find some untranslated data for unknown reasons and drop out them to get the last edition of translated STS datasets. We also do the same example on this pruned data, the performance is shown in the \textit{Last edition} of Table \ref{tab1}. Table \ref{tab2} shows the statistics of our Japanese STS benchmark. We believe the last edition is valid compared to the other machine-translated datasets.

\begin{table}[]
\centering
\begin{adjustbox}{width=\textwidth}
\begin{tabular}{llccccccc}
\hline \hline
\multicolumn{2}{l}{\textbf{Google translator}}  &        &        &        &        &        &        &        \\ \hline
\multicolumn{2}{l|}{BLEU   threshold}  & -      & 0.05   & 0.1    & 0.15   & 0.2    & 0.25   & 0.3    \\ \hline
datasize     & \multicolumn{1}{l|}{}   & 21172  & 20612  & 20449  & 19508  & 18207  & 16694  & 14904  \\ \hline
MAE          & \multicolumn{1}{l|}{}   & 0.4415 & 0.4443 & 0.4807 & 0.4493 & 0.4963 & 0.4966 & 0.4916 \\ \hline
MSE          & \multicolumn{1}{l|}{}   & 0.3384 & 0.336  & 0.3793 & 0.3332 & 0.412  & 0.4191 & 0.4018 \\ \hline
R2           & \multicolumn{1}{l|}{}   & 0.6481 & 0.6505 & 0.6058 & 0.6537 & 0.5727 & 0.5632 & 0.5822 \\ \hline\hline
\multicolumn{2}{l}{\textbf{DeepL   translator}} &        &        &        &        &        &        &        \\ \hline
\multicolumn{2}{l|}{BLEU   threshold}  & -      & 0.05   & 0.1    & 0.15   & 0.2    & 0.25   & 0.3    \\ \hline
datasize     & \multicolumn{1}{l|}{}   & \textbf{21172}  & 19337  & 19144  & 18368  & 17535  & 16683  & 15727  \\ \hline
MAE          & \multicolumn{1}{l|}{}   &\textbf{0.4206} & 0.433  & 0.4314 & 0.4397 & 0.451  & 0.4448 & 0.442  \\ \hline
MSE          & \multicolumn{1}{l|}{}   &\textbf{0.3058} & 0.3135 & 0.3117 & 0.324  & 0.342  & 0.3312 & 0.328  \\ \hline
R2           & \multicolumn{1}{l|}{}   & \textbf{0.6894} & 0.6802 & 0.6834 & 0.6711 & 0.6529 & 0.6638 & 0.6666 \\ \hline\hline
\multicolumn{2}{l}{\textbf{Last edition}} &        &        &        &        &        &        &        \\ \hline
datasize     &       &  19330 & MAE    & \textbf{0.4119} & MSE    & \textbf{0.2821} & R2     & \textbf{0.7137} \\ \hline\hline
\end{tabular}
\end{adjustbox}
\caption{The performance of different models trained on different translated data. We use Mean Absolute Error (MAE), Mean Square Error (MSE), and R-Squared (R2) as metrics to evaluate translation qualities. The model trained on the last edition translated STS dataset has the best performance on MAE, MSE and R2 metrics among any other models.}
\label{tab1}
\end{table}

\begin{table}[]
\centering
\begin{tabular}{l|c|c}
\hline
Japanese   STS benchmark          & Translated by & \#Examples \\ \hline
\textbf{Total}                    & -             & 43165      \\
\textbf{STS12}                    & Machine       & 2615       \\
\textbf{STS13}                    & Machine       & 1063       \\
\textbf{STS14}                    & Machine       & 3165       \\
\textbf{STS15}                    & Machine       & 2901       \\
\textbf{STS16}                    & Machine       & 1143       \\
\textbf{STS-B}                    & Machine       & 8443       \\
\textbf{JSICK}                    & Human         & 9927       \\
\textbf{JSTS}                     & Human         & 13908      \\ \hline
\end{tabular}
\caption{Statistics of Japanese STS benchmark.}
\label{tab2}
\end{table}

\subsection{STS Results}

We compare existing sentence representation methods on the Japanese STS benchmark defined in the previous section. The purpose of this experiment is to examine the effectiveness of existing sentence representation methods and choose the ones that will be compared to JCSE.

Because there are no available pre-trained Japanese sentence embedding models or we do not know the training details of the public checkpoints, we need to train these models and report the results by ourselves. We consider both unsupervised and supervised sentence representation models to apply a comprehensive evaluation. Unsupervised baselines include average FastText embeddings \cite{ref56}, average BERT embeddings \cite{ref25}, and post-processing methods such as BERT-flow \cite{ref57} and BERT-whitening \cite{ref58}. We also compare several recent state-of-the-art contrastive learning methods including SimCSE \cite{ref7}, MixCSE \cite{ref13}, and DiffCSE \cite{ref14}. Other supervised methods include SBERT \cite{ref59}, CoSENT\footnote[9]{https://github.com/bojone/CoSENT} and SimCSE \cite{ref7}.

We apply official codes to train a Japanese sentence embedding model based on each of the above-mentioned sentence representation methods. We tried various hyperparameters in the training process, and find that the officially recommended hyperparameters are better than others in most situations. Thus, in this section, all the Japanese sentence embedding models are trained in the hyperparameters referred to as relative officially recommended training settings. The models trained based on BERT-base or BERT-large using their public checkpoints\footnote[10]{https://huggingface.co/cl-tohoku/bert-base-japanese \& https://huggingface.co/cl-tohoku/bert-large-japanese}.

For unsupervised contrastive learning methods, we apply the Wikipedia data as training data. We use WikiExtractor\footnote[11]{https://github.com/attardi/wikiextractor} and GiNZA to make unlabeled sentences from the latest Japanese Wikipedia corpus, and randomly sample $10^{6}$ sentences as training data. For supervised methods, all the models are trained just on the JSNLI dataset due to data availability.

Table \ref{tab3} shows the evaluation results. The performances of Japanese sentence embedding models on the Japanese STS tasks is a little different from those on the English STS benchmark tasks. For the unsupervised methods, recent state-of-the-art contrastive learning methods perform better than previous baselines, and the MixCSE method shows the best or nearly the best results on almost all STS tasks. Surprisingly, we find that models trained on RoBERTa \cite{ref60} pre-trained checkpoints\footnote[12]{https://huggingface.co/nlp-waseda/roberta-base-japanese \& https://huggingface.co/nlp-waseda/roberta-large-japanese} using a contrastive learning method such as SimCSE (i.e., SimCSE-RoBERTa-base model) get poor results even worse than previous baselines on some STS tasks, so we do not choose models trained on RoBERTa checkpoints to compare with JCSE. For supervised methods, the models using the contrastive learning method should achieve better performance than the previous state-of-the-art SBERT \cite{ref59} model, reversely, SBERT beyond SimCSE in our experiments because of much smaller training data compared to those in the English experiments. Specifically, we use the JSNLI dataset which is about half of the training data in the English experiments \cite{ref7}, which uses the SNLI \cite{ref11} and MNLI \cite{ref12} datasets. This point was easily overlooked in other research but uncovered in our experiment. This is also the reason that unsupervised contrastive learning methods get better performance than supervised contrastive learning methods in our experiments. After checking Table \ref{tab3}, we can find that the results on the human-annotated STS datasets and the machine-translated datasets indicate different trends. This reflects the obvious gap between machine translation and human annotation. To emphasize the performance on the human-annotated STS tasks, we also show the results just on JSICK and JSTS datasets in Table \ref{tab4}. The results show that supervised contrastive learning methods get nearly competitive performance compared with unsupervised contrastive learning methods even the labeled training data is limited. Contrastive learning methods surpass previous methods which are similar to the results in Table \ref{tab3}.

\begin{table}[]
\centering
\begin{adjustbox}{width=\textwidth}
  \begin{tabular}{lccccccccc}
\hline
\textbf{Model}              & \textbf{STS12} & \textbf{STS13} & \textbf{STS14} & \textbf{STS15} & \textbf{STS16} & \textbf{STS-B} & \textbf{JSICK} & \textbf{JSTS} & \textbf{Avg.} \\ \hline
\multicolumn{10}{c}{\textit{Unsupervised models}}                                                                                                                                  \\ \hline
FastText embeddings (avg.)  & 42.78          & 51.71          & 42.45          & 54.24          & 54.06          & 51.30          & 74.76          & 60.23         & 53.94         \\
BERT-base (first-last avg.) & 38.66          & 52.06          & 36.41          & 55.69          & 57.06          & 51.39          & 71.97          & 65.19         & 53.55         \\
BERT-base-flow              & 44.25          & 55.60          & 42.62          & 65.77          & 60.67          & 55.10          & 77.26          & 64.90         & 58.27         \\
BERT-base-whitening         & 39.14          & 55.19          & 41.76          & 57.27          & 57.93          & 46.26          & 71.86          & 56.00         & 53.18         \\ \hline
SimCSE-BERT-base            & 56.82          & 66.45          & 55.66          & \textbf{74.80}          & 72.73          & 69.71          & 77.20          & 71.66         & 68.13         \\
SimCSE-BERT-large           & 55.26          & 62.66          & 51.95          & 72.76          & 69.49          & 65.20          & 77.63          & 72.98         & 65.99         \\
SimCSE-RoBERTa-base         & 48.65          & 60.09          & 46.97          & 66.10          & 65.29          & 59.88          & 72.32          & 63.82         & 60.39         \\
MixCSE-BERT-base            & 55.07          & \textbf{67.45}          & \textbf{56.94}          & 74.35          & \textbf{73.11}          & \textbf{70.21}          & 76.97          & 71.85         & \textbf{68.24}         \\
MixCSE-BERT-large           & 56.56          & 64.59          & 54.84          & 74.55          & 70.85          & 67.76          & \textbf{78.15}          & \textbf{73.89}        & 67.65         \\
DiffCSE-BERT-base           & \textbf{57.59}          & 67.07          & 55.93          & 74.07          & 72.88          & 69.37          & 76.29          & 71.07         & 68.03         \\ \hline
\multicolumn{10}{c}{\textit{Supervised models}}                                                                                                                                    \\ \hline
SBERT-base                  & 58.79          & 60.51          & 53.30          & 68.64          & \textbf{66.59}         & 64.76          & 65.25          & 75.28         & 64.14         \\
SBERT-large                 & 60.57          & \textbf{64.27}          & \textbf{54.24}          & \textbf{70.92}          & 65.91          & \textbf{66.17}          & 67.68          & \textbf{76.84}         & \textbf{65.83}         \\
CoSENT                      & \textbf{61.73}          & 58.28          & 52.46          & 64.72          & 63.02          & 63.27          & 56.81          & 74.17         & 61.81         \\
SimCSE-BERT-base            & 51.65          & 54.43          & 46.49          & 63.92          & 60.81          & 57.33          & 74.19          & 69.12         & 59.74         \\
SimCSE-BERT-large           & 50.29          & 56.47          & 48.10          & 67.76          & 59.64          & 59.63          & \textbf{77.48}          & 69.73         & 60.80         \\ \hline
  \end{tabular}
\end{adjustbox}
\caption{Sentence embedding performance on STS tasks (Spearman's correlation, ``all'' setting). For BERT-flow and BERT-whitening we only report the ``NLI'' setting.}
\label{tab3}
\end{table}

\begin{table}[]
\centering
\begin{tabular}{lccc}
\hline
\textbf{Model}              & \textbf{JSICK} & \textbf{JSTS}  & \textbf{Avg.}  \\ \hline
\multicolumn{4}{c}{\textit{Unsupervised models}}                               \\ \hline
FastText embeddings (avg.)  & 74.76          & 60.23          & 67.50          \\
BERT-base (first-last avg.) & 71.97          & \textbf{65.19} & 68.58          \\
BERT-base-flow              & \textbf{77.26} & 64.90          & \textbf{71.08} \\
BERT-base-whitening         & 71.86          & 56.00          & 63.93          \\ \hline
SimCSE-BERT-base            & 77.20          & 71.66          & 74.43          \\
SimCSE-BERT-large           & 77.63          & 72.98          & 75.31          \\
SimCSE-RoBERTa-base         & 72.32          & 63.82          & 68.07          \\
MixCSE-BERT-base            & 76.97          & 71.85          & 74.41          \\
MixCSE-BERT-large           & \textbf{78.15} & \textbf{73.89} & \textbf{76.02} \\
DiffCSE-BERT-base           & 76.29          & 71.07          & 73.68          \\ \hline
\multicolumn{4}{c}{\textit{Supervised models}}                                 \\ \hline
SBERT-base                  & 65.25          & 75.28          & 70.27          \\
SBERT-large                 & 67.68          & \textbf{76.84} & 72.26          \\
CoSENT                      & 56.81          & 74.17          & 65.49          \\
SimCSE-BERT-base            & 74.19          & 69.12          & 71.66          \\
SimCSE-BERT-large           & \textbf{77.48} & 69.73          & \textbf{73.61} \\ \hline
\end{tabular}
\caption{Sentence embedding performance on human-annotated STS tasks (Spearman's correlation, ``all'' setting).}
\label{tab4}
\end{table}

\subsection{Relevant Content Words}

The core insight of JCSE is to generate target domain contradiction sentence pairs for contrastive learning. As we described above, we find that nouns are the most relevant content words and construct negative pairs for the anchor sentences by substituting noun phrases in them. Empirically, not all word types play an equal role in determining the semantics of a Japanese sentence. A previous study~\cite{ref23} insufficiently verifies this opinion but gives us inspiration.

% in Japanese sentences for multiple Japanese language models based on our experiments. It means that we can 

In this section, we investigate which part-of-speech (POS) tags mainly influence on examining if a Japanese sentence pair is semantically similar or not for different Japanese sentence embedding models. To measure this, we borrow the approach used in TSDAE~\cite{ref27}. We select a sentence pair $(a,b)$ that is labeled as relevant and find the word that maximally reduces the cosine-similarity score for the pair $(a,b)$:

\begin{equation}\label{eq6}
    \hat{w} = \mathrm{argmax} _{w} (\mathrm{cossim}(a,b)-\mathrm{min}(\mathrm{cossim}(a\setminus w,b),\mathrm{cossim}(a,b\setminus w)))
\end{equation}
among all words $w$ that appear in either a or b. We record the POS tags for $\hat{w}$ and compute the distribution of POS tags among all sentence pairs. The POS tags are defined using spaCy \cite{ref54}.

\begin{figure}[]
    \centering
    \includegraphics[width=\textwidth]{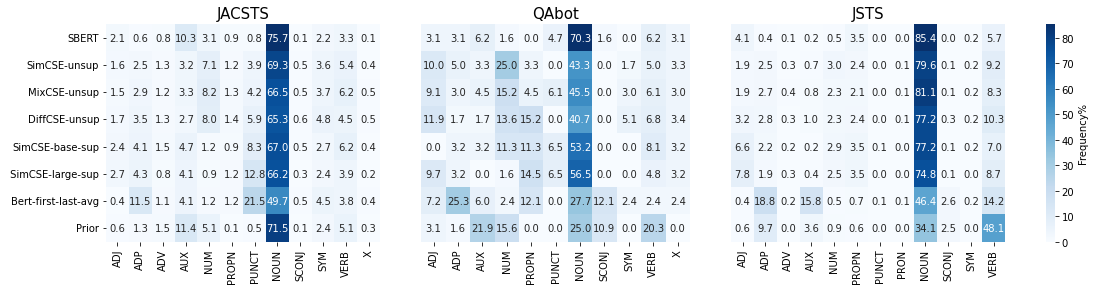}
    \caption{POS tags for the most relevant content words in sentences. All POS tags are defined by spaCy. X represents other. The most relevant content word in a sentence that mostly influences if a sentence pair is considered as similar.}
    \label{fig2}
\end{figure}

The results on the three datasets are shown in Figure \ref{fig2}. We find that nouns (NOUN) are by far the most relevant content words in a Japanese sentence across the three datasets, and we do not perceive significant differences between all three datasets. This result is important for our method because it experimentally validates the theoretical foundation of our method. That is, it is practical to substitute nouns to construct contradiction sentence pairs.

% This is good news for our method. The relevant content words experiments show the theory fundamental for our method that it is practical 

\subsection{Experiment Setting}
In this section, we describe two experiment settings in our method, including data generation, and domain adaptation.

\textbf{Data generation}: We bridge a gap between the scarcity of domain-specific data in low-resource language and the demand of sufficient training data for contrastive learning using data generation. In our work, we first finetune the pre-trained T5-base checkpoint on the target domain corpus to get our target domain data generator. Then we mask all noun chunks in every sentence with unique sentinel tokens as an input of the data generator. Afterward, we apply a simple unsampled beam search decode algorithm \cite{ref61,ref62} implemented in the Hugging Face \cite{ref63} to generate target domain data. Our approach automatically generates data in the target domain. Real examples are shown in Table \ref{tab5}. For the finetune process, we initialize the T5 model from public checkpoint\footnote[12]{https://huggingface.co/sonoisa/t5-base-japanese-v1.1}. We use Adafactor \cite{ref64} as the optimizer and set the learning rate to 0.001 implemented using Pytorch and Huggingface and trained on three NVIDIA RTX 2080Ti.

\begin{CJK}{UTF8}{min} 
\begin{table}[]
\centering
\begin{adjustbox}{width=\textwidth}
\begin{tabular}{lllllllllllllll}
\textbf{Clinical Domain}            &            &            &           &          &          &         &         &        &        &        &        &        &        &  \\ \cline{1-14}
\multicolumn{8}{l}{\textbf{Original Sentence}}                                                                      &        &        &        &        &        &        &  \\
\multicolumn{8}{l}{幻聴が増悪し、アルコール性精神障害の合併が疑われ、精神科受診が適切と判断された。}                                                        &        &        &        &        &        &        &  \\
\multicolumn{14}{l}{The auditory hallucinations worsened, complications of alcoholic psychosis were suspected, and a psychiatric consultation was deemed appropriate.}     &  \\
\multicolumn{8}{l}{\textbf{Synthetic Contradiction Sentence}}                                                       &        &        &        &        &        &        &  \\
\multicolumn{9}{l}{徐々に症状が増悪し、急性大動脈解離の急性大動脈解離が疑われ、緊急手術が適切と判断された。}                                                             &        &        &        &        &        &  \\
\multicolumn{15}{l}{Gradual worsening   of symptoms, acute aortic dissection of acute aortic dissection was   suspected, and emergency surgery was deemed appropriate.}      \\
\multicolumn{15}{l}{アルコール依存症の症状が増悪し、急性大動脈解離のリスクファクターの増大が疑われ、緊急手術が適切と判断された。}                                                                                                  \\
\multicolumn{14}{l}{The patient's   symptoms of alcoholism were exacerbated and an increased risk factor for   acute aortic dissection was suspected, and}                &  \\
\multicolumn{4}{l}{emergency   surgery was deemed appropriate.}           &          &          &         &         &        &        &        &        &        &        &  \\
\multicolumn{14}{l}{少しずつ症状が増悪し、カテコラミン心筋症の再発や悪化が疑われ、精神科の受診が適切と判断された。}                                                                                                      &  \\
\multicolumn{14}{l}{The symptoms   gradually worsened, and a psychiatric consultation was deemed appropriate due   to a suspected recurrence or}                          &  \\
\multicolumn{4}{l}{worsening   of catecholaminergic cardiomyopathy.}      &          &          &         &         &        &        &        &        &        &        &  \\
\multicolumn{12}{l}{徐々に意識障害が増悪し、急性硬膜外血腫の再発や急性大動脈解離が疑われ、脳底動脈造影CT撮影が適切と判断された。}                                                                            &        &        &  \\
\multicolumn{14}{l}{Gradually worsening   loss of consciousness, recurrent acute epidural hematoma and acute aortic   dissection were suspected, and CT imaging with}     &  \\
\multicolumn{7}{l}{contrast-enhanced   cerebral basilar artery angiography was deemed appropriate.}       &         &        &        &        &        &        &        &  \\
\multicolumn{12}{l}{来院後徐々に症状が増悪し、抗ヒスタゾールや精神科受診の妥当性も低いことが疑われ、精神科受診が適切と判断された。}                                                                            &        &        &  \\
\multicolumn{14}{l}{Symptoms gradually   worsened after the patient's visit, and it was suspected that antihistazol   and psychiatric consultation were not appropriate.} &  \\
\multicolumn{12}{l}{症例は徐々に症状が増悪し、COVID-19の急性増悪のリスクファクターの増悪が疑われ、COVID-19の診断が適切と判断された。}                                                                   &        &        &  \\
\multicolumn{14}{l}{The patient had a   gradual exacerbation of symptoms, and a diagnosis of COVID-19 was deemed   appropriate due to the suspected exacerbation of}      &  \\
\multicolumn{4}{l}{risk factors   for acute exacerbation of COVID-19.}    &          &          &         &         &        &        &        &        &        &        &  \\ 
       &            &            &           &          &          &         &         &        &        &        &        &        &        &  \\
\textbf{Educational Domain}                &            &            &           &          &          &         &         &        &        &        &        &        &        &  \\ \cline{1-14}
\textbf{Original Sentence}                  &            &            &           &          &          &         &         &        &        &        &        &        &        &  \\
\multicolumn{14}{l}{とあるクラスで作成されたオブジェクトを他クラスで操作するためにはどのようにすれば良いでしょうか?}                                                                                                     &  \\
\multicolumn{14}{l}{How can an object   created in one class be manipulated by another class?}                                                                            &  \\
\multicolumn{3}{l}{\textbf{Synthetic   Contradiction Sentence}}        &           &          &          &         &         &        &        &        &        &        &        &  \\
\multicolumn{14}{l}{MiniDuckSimulator.javaで作成されたオブジェクトをSystem.javaで操作するためにはどのようにすれば良いでしょうか?}                                                                              &  \\
\multicolumn{14}{l}{How can I use   System.java to manipulate objects created in MiniDuckSimulator.java?}                                                                 &  \\
\multicolumn{14}{l}{AdderServletクラスで作成されたオブジェクトをServletクラスで操作するためにはどのようにすれば良いでしょうか?}                                                                                      &  \\
\multicolumn{13}{l}{How can I   manipulate objects created in the AdderServlet class with the Servlet class?}                                                    &        &  \\
\multicolumn{13}{l}{MetroFair.javaで作成されたオブジェクトをSafariの環境で操作するためにはどのようにすれば良いでしょうか?}                                                                               &        &  \\
\multicolumn{13}{l}{How can I   manipulate objects created in MetroFair.java in the Safari environment?}                                                         &        &  \\
\multicolumn{13}{l}{MallardDuckクラスで作成されたオブジェクトをSystem.javaで操作するためにはどのようにすれば良いでしょうか?}                                                                             &        &  \\
\multicolumn{13}{l}{How can I use   System.java to manipulate objects created in the MallardDuck class?}                                                         &        &  \\
\multicolumn{13}{l}{DecoyDuckクラスで作成されたオブジェクトをGameDuckクラスで操作するためにはどのようにすれば良いでしょうか?}                                                                               &        &  \\
\multicolumn{13}{l}{How can an object   created in the DecoyDuck class be manipulated in the GameDuck class?}                                                    &        &  \\
\multicolumn{13}{l}{DecoyDuckSimulator.javaで作成されたオブジェクトをGecoyDuckSimulator.javaで操作するためにはどのようにすれば良いでしょうか?}                                                        &        &  \\
\multicolumn{14}{l}{How can I use   GecoyDuckSimulator.java to manipulate objects created in   DecoyDuckSimulator.java?}                                                  &
\end{tabular}
\end{adjustbox}
\caption{Examples of domain-targeted synthetic contradiction sentence pairs based on original sentences. As we described, we apply an unsampled beam search decode algorithm to generate sentences, define different \textit{num} \textit{return} \textit{sequences} in the huggingface to get different size of generated sentences.} 
\label{tab5}
\end{table}
\end{CJK}

\textbf{Domain adaption}: JCSE focuses on domain adaption characterized by the following two-stage contrastive learning recipe. A pre-trained sentence embedding model is firstly adapted to a target domain by fine-tuning it on the synthesized generated sentence pairs. Then, the model is further fine-tuned on labeled JSNLI data. We find that this training sequence, first on the synthesized generated data and then on JSNLI data, is better for our method where reverse training sequence can harm the performance.

% In this case, we further train the Japanese language models on target domain data from our synthesized generated data and general domain data from the JSNLI dataset. Our target domain Japanese sentence embedding models are trained in a two-stage contrastive learning recipe, first adapting the sentence embedding to a specific domain by finetuning the pre-trained language models on the synthesized generated sentence pairs, then finetuning on the labeled JSNLI data. 

In our experiment, both two training stages utilize the improved contrastive learning method that incorporates different weighting with different hard negatives. The training objective is described in section 3.3. In stage one, to formulate the triple training data $\left \{ v,v^{+},v^{-}  \right \}$, we apply a generated contradictory sentence as $v^{-}$ based on $v$ to get a negative pair. For the entailment sentence pairs, we simply feed the sentence $v$ to the encoder twice and get two different embeddings with different dropout masks $v$, $v^{+}$. In stage two, we construct triple training data using the JSNLI dataset that choose the sentence pair whose label is entailment as the positive pair $v$, $v^{+}$ and find the negative sentence $v^{-}$ in the pair whose label is contradiction for the anchor sentence $v$.

We build JCSE upon BERT\cite{ref25} models\footnote[13]{https://huggingface.co/cl-tohoku/bert-base-japanese \& https://huggingface.co/cl-tohoku/bert-large-japanese}. Our models are implemented using Pytorch and Huggingface, trained on a single NVIDIA TITAN RTX GPU. During stage one in the training, we set the learning rate to 5e-5 for the \textit{base} model and 1e-5 for the \textit{large} model. We use a batch size of 512 for all models and set the temperature $\tau$ and the weighting $\alpha$ to $0.05$ and $0$, respectively. During stage two, we set $\alpha$ to $1$, other training settings are the same as in stage one. Regarding the training data scale in stage one, we find that the best performance for the \textit{base} model is obtained by generating 4 contradictory sentences for every anchor sentence, meaning that 4 times more triplets are formed compared to the size of training data. For the \textit{large} model, 6 times generated contradictory sentences are applied to get the best performance.

\subsection{Domian-specific downstream tasks}

We evaluate JCSE on two Japanese domain-specific downstream tasks, STS in a clinical domain and information retrieval in an educational domain. The former and latter defined by JACSTS~\cite{ref29} and QAbot datasets, respectively. Note that these two datasets are human-annotated and our method uses no task-specific labeled data, instead it generates contradictory sentences for original sentences and forms triplets to fine-tune a pre-trained model.

% in our method, we do not use any task-specific training data.

\textbf{JACSTS} JACSTS dataset~\cite{ref29} is a publicly available dataset\footnote[14]{https://github.com/sociocom/Japanese-Clinical-STS}  for sentence-level clinical STS from Japanese case reports. The dataset consists of 3670 sentence pairs annotated on a scale from $0$ to $5$, where $0$ means that the two sentences are semantically very dissimilar, and $5$ means that they are very similar.

\textbf{QAbot} QAbot dataset is collected from realistic question-answer dialogues in certain courses by a bot developed on Slack. The dataset consists of 1142 question-answer pairs that are individually annotated with relevance to each of the 20 queries. For these 20 queries, there is at least one relative question labeled as 1 in the dataset. We publicly share this private dataset for this work\footnote[15]{https://github.com/mu-kindai/JCSE}.

For the JACSTS dataset, because it is an STS task, we combine it into the SentEval toolkit and measure the spearman score as the result. For the QAbot dataset, it is a question retrieval task. Hence, we apply a simple first-stage dense retrieval system with two-tower architecture by referring to previous studies \cite{ref43,ref46,ref65}. More specifically, we extract the mean embedding on top of layers to obtain the final query/question representation from two-tower retrieval models consisting of two independent Japanese sentence embedding models: one for each of the query and question towers, respectively. We use Mean Average Precision (MAP), Mean Reciprocal Rank (MRR), and Precision at N (P@N) as the evaluation metrics.

During the evaluation process, we first consider the direct transfer scenario where generated target domain data are not used for sentence embedding training. Then we evaluate the model for domain adaption, where synthesized generated target domain data are used in stage one, and the general domain JSNLI data are used in stage two. We also combine the unlabeled target domain corpus with the Wikipedia corpus for the unsupervised contrastive learning method in the training process. The results for the two tasks are shown in Table \ref{tab6}.

\begin{table}[]
\centering
\begin{adjustbox}{width=\textwidth}
\begin{tabular}{lcccccc}
\hline
\textbf{Model}              & \textbf{Fine-tune Data} &  \textbf{JACSTS}&\multicolumn{4}{c}{\textbf{QAbot}}                                \\ \cline{4-7}
                            &               &          & \textbf{MRR}    & \textbf{MAP}    & \textbf{P@1}  & \textbf{P@5}  \\ \hline
\multicolumn{6}{c}{\textit{Direct transfer}}                                                                              \\ \hline
FastText embedding (avg.)   & N/A        & 75.22             & 0.6330          & 0.5190          & 0.55          & 0.24          \\
BERT-base (first-last avg.) & N/A       & \textbf{79.31}              & 0.6023          & 0.5150          & 0.55          & 0.22          \\ \hline
SimCSE-BERT-base            & Wiki       & \textbf{81.43}             & \textbf{0.7690}          & \textbf{0.6384}          & \textbf{0.70}          & 0.31          \\
MixCSE-BERT-base            & Wiki       & 80.96             & 0.7591          & 0.6261          & 0.70          & \textbf{0.32}          \\
DiffCSE-BERT-base           & Wiki       & 78.17              & 0.7259          & 0.6179          & 0.65          & 0.32          \\ \hline
SBERT-base                  & JSNLI       & 75.69              & 0.5922          & 0.4080          & 0.55          & 0.21          \\
SBERT-large                 & JSNLI      & 73.77              & 0.5263          & 0.3475          & 0.45          & 0.19          \\
SimCSE-BERT-base            & JSNLI     & 79.86               & 0.7544          & 0.5459          & 0.70          & 0.26          \\
SimCSE-BERT-large           & JSNLI     & \textbf{81.64}                & \textbf{0.7666} & \textbf{0.6093} & \textbf{0.70} & \textbf{0.31} \\ \hline
\multicolumn{6}{c}{\textit{Domain Adaptation}}                                                                   \\ \hline
SimCSE-BERT-base            & Wiki\&In-domain    & 81.69     & 0.7632          & 0.6366          & 0.70          & 0.32          \\
MixCSE-BERT-base            & Wiki\&In-domain   & \textbf{82.99}      & 0.7707          & 0.6186          & 0.70          & 0.32          \\
DiffCSE-BERT-base           & Wiki\&In-domain    & 81.20       & \textbf{0.7796} & \textbf{0.6489} & \textbf{0.70} & \textbf{0.33} \\ \hline
JCSE-base                   & In-domain    & 81.80            & 0.7576          & 0.6184          & 0.70          & 0.31          \\
JCSE-large                  & In-domain     & 81.16          & 0.7978          & 0.6633          & 0.75          & 0.33          \\
JCSE-base                   & In-domain$\rightarrow$JSNLI   & \textbf{82.08}      & 0.7820          & 0.6211          & 0.75          & 0.29          \\
JCSE-large                  & In-domain$\rightarrow$JSNLI    & \textbf{82.43}       & \textbf{0.8173} & \textbf{0.6812} & \textbf{0.75} & \textbf{0.35} \\ \hline
\end{tabular}
\end{adjustbox}
\caption{Performance on two domain-specific downstream tasks.}
\label{tab6}
\end{table}

Table~\ref{tab6} shows that our method improves the performance through domain adaptation using synthesized generated in-domain data. Even just using the generated data after the first stage of contrastive learning training can achieve competitive results compared with the out-of-the-box sentence embedding models for direct transfer which means that the first stage training uses much less training data than the data used in the direct transfer but gets improved performance. And further training using the JSNLI dataset can further improve our method's performance after the second stage of contrastive learning training. Superingsly, we also find that combining the in-domain corpus with the Wikipedia corpus in the training process can benefit existing unsupervised contrastive learning methods. The significant and consistent improvements over the two tasks also demonstrate the effectiveness of our data generation approach by fine-tuning a T5 model. Furthermore, compared to direct transfer and just directly integrating the in-domain corpus into existing contrastive learning methods, our approach achieves the best performance on almost all metrics with less training data scale, which convincingly demonstrate that JCSE is an efficient and effective Japanese sentence representation framework for better sentence embedding on Japanese domain-specific downstream tasks.

\section{Conclusion and Future Work}

In this paper, we proposed a novel effective Japanese sentence representation learning framework, JCSE, for the Japanese domain-specific downstream tasks. Meanwhile, we establish a Japanese STS benchmark to evaluate various Japanese sentence embedding models using previous state-of-the-art sentence representation learning methods.

% and evaluate their performance on our proposed Japanese STS benchmark.

We first collect a target domain corpus and finetune a pre-trained data generator (T5 model) on this unlabeled data to get our target domain data generator. Then, based on our relevant content words experiments, we substitute all nouns in sentences by the data generator to generate contradictory sentence pairs. Afterward, we use the synthesized target domain training data with contrastive learning to train a sentence embedding model. We evaluate the models on two domain-specific downstream tasks, one of which uses the QAbot dataset originally collected by us. Our method achieves significantly better performance after the domain adaptation training, compared with direct transfer and current state-of-the-arts in Japanese. We believe that JCSE paves a practicable way for domain-specific downstream tasks in low-resource languages. 

Our future work includes the establishment of a new Japanese STS benchmark based on human annotation, in contrast to the current benchmark based on machine translation. In addition, we plan to explore a better-synthesizing strategy to generate better sentence pairs and evaluate our method's performance on more diverse target domains.

%%\label{}





\end{document}